\documentclass[10pt,twocolumn,letterpaper]{article}

\usepackage{cvpr}              

\usepackage{amsmath,amssymb}
\usepackage{bm}        
\usepackage{bbm}       
\usepackage{algorithm}
\usepackage{algorithmic}
\usepackage{enumitem} 
\usepackage{multirow} 
\usepackage{xcolor}
\usepackage[table]{xcolor}
\usepackage{amsmath}
\usepackage{amsfonts}
\usepackage{bm}   
\usepackage{makecell}

\definecolor{cvprblue}{rgb}{0.21,0.49,0.74}
\usepackage[pagebackref,breaklinks,colorlinks,allcolors=cvprblue]{hyperref}

\def\paperID{14538} 
\def\confName{CVPR}
\def\confYear{2026}

\title{FairMT: Fairness for Heterogeneous Multi-Task Learning}

\author{Guanyu	Hu\\	
Xi'an Jiaotong University\\
Queen Mary University of London
\and 
Tangzheng Lian\\
Kings College London
\and 
Na Yan\\
Kings College London
\and
Dimitrios Kollias\\	
Queen Mary University of London
\and
Xinyu Yang\\
Xi'an Jiaotong University
\and
Oya Celiktutan\\
Kings College London
\and
Siyang Song\\
University of Exeter
\and
Zeyu Fu\\
University of Exeter
}

\begin{document}
\maketitle
\begin{abstract}
Fairness in machine learning has been extensively studied in single-task settings, while fair multi-task learning (MTL), especially with heterogeneous tasks (classification, detection, regression) and partially missing labels, remains largely unexplored. Existing fairness methods are predominantly classification-oriented and fail to extend to continuous outputs, making a unified fairness objective difficult to formulate. Further, existing MTL optimization  is structurally misaligned with fairness: constraining only the shared representation, allowing task heads to absorb bias and leading to uncontrolled task-specific disparities. Finally, most work treats fairness as a zero-sum trade-off with utility, enforcing symmetric constraints that achieve parity by degrading well-served groups. We introduce \emph{\textsc{FairMT}}, a unified fairness-aware MTL framework that accommodates all three task types under incomplete supervision. At its core is an Asymmetric Heterogeneous Fairness Constraint Aggregation mechanism, which consolidates task-dependent asymmetric violations into a unified fairness constraint. Utility and fairness are jointly optimized via a primal--dual formulation, while a head-aware multi-objective optimization proxy provides a tractable descent geometry that explicitly accounts for head-induced anisotropy. Across three homogeneous and heterogeneous MTL benchmarks encompassing diverse modalities and supervision regimes, \emph{\textsc{FairMT}} consistently achieves substantial fairness gains while maintaining superior task utility. Code will be released upon paper acceptance.

\end{abstract}    

\begin{figure*}[htbp]
\centering
\includegraphics[width=1\linewidth]{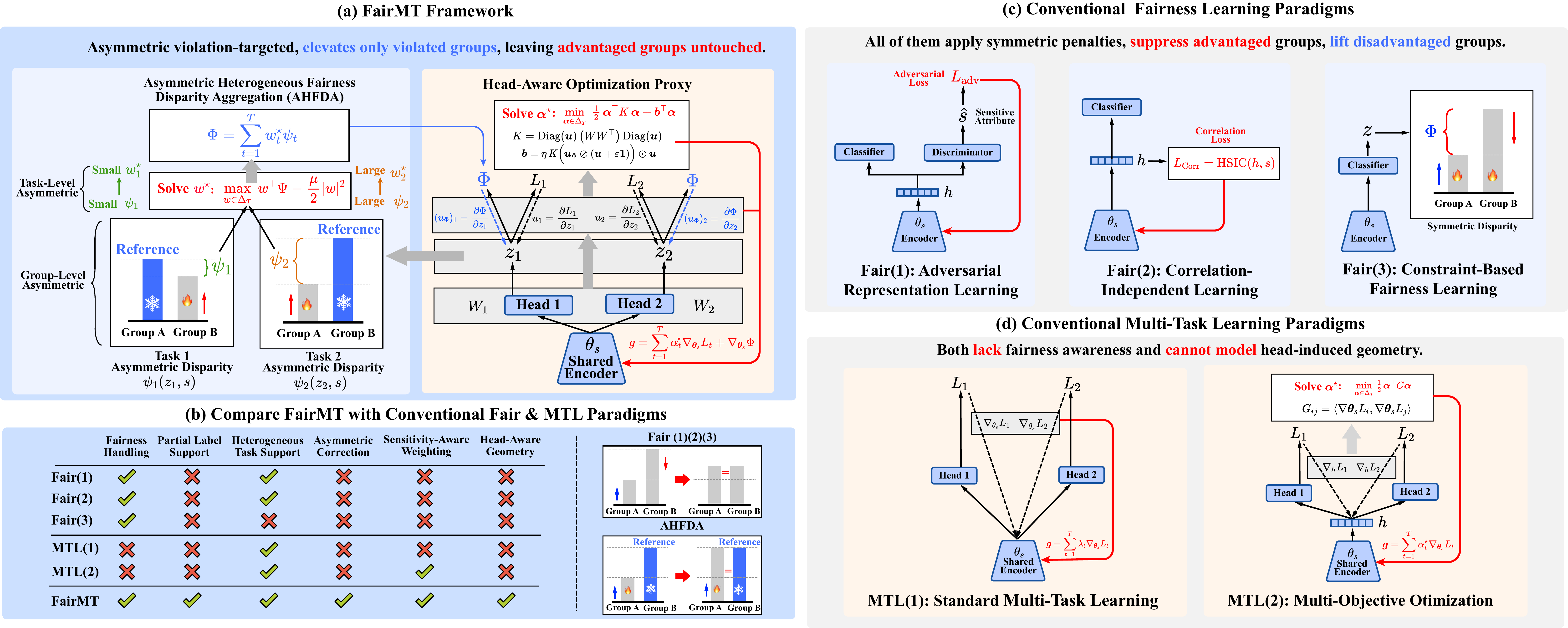}
\caption{
{\small
\textbf{Comparison of FairMT with Conventional Fair or MTL Paradigms.}
(a) Our proposed FairMT framework.
(b) Comparison between FairMT and conventional paradigms: the left panel highlights our advantages, and the right panel illustrates how AHFDA achieves fairness by asymmetrically lifting only violated groups without harming advantaged groups.
(c) Conventional fairness learning paradigms, all applying symmetric penalties that simultaneously suppress advantaged groups and lift disadvantaged groups.
(d) Conventional multi-task learning paradigms, both lacking fairness awareness and unable to model head-induced geometry.}
\vspace{-1em}
}
\label{fig:framework}
\end{figure*}
\vspace{-1em}

\section{Introduction}
\label{sec:intro}
Fairness has become a fundamental requirement for machine learning systems deployed in socially sensitive domains such as healthcare, hiring, and credit lending \cite{mehrabi2021survey,pessach2022review,soremekun2022software}. 
However, research on algorithmic fairness has focused almost exclusively on the single-task setting \cite{zafar2017fairness,menon2018cost,zafar2019fairness}, primarily in classification contexts. 
In contrast, fairness in \emph{multi-task learning} (MTL) remains largely unexplored. 
The coexistence of multiple and often heterogeneous tasks makes it difficult to define and enforce fairness consistently across objectives, resulting in conflicting criteria and incompatible optimization goals. 
Thus, directly applying conventional fairness methods leads to degraded performance.

Achieving fairness in MTL is impeded by several challenges:
\textbf{(1) Fairness by Suppression:}
Prevailing fairness paradigms (\textbf{Fig.~\ref{fig:framework}(c)}) can be broadly categorized into (i) adversarial representation learning\cite{madras2018learning,zhang2018mitigating}, (ii) correlation-independent regularization (e.g., HSIC fairness) \cite{dehdashtian2024utility,quadrianto2019discovering}, and (iii) constraint-based optimization \cite{zafar2019fairness,agarwal2018reductions}, they address bias largely through suppression rather than enhancement.
first two Representation-based approaches enforce fairness by decorrelating learned features from sensitive attributes, but this often induces representational collapse and weakens task-relevant semantics, degrading performance across all groups.
Constraint-based formulations directly penalize disparity gaps yet tend to equalize outcomes by lowering the performance of well-served groups instead of elevating disadvantaged ones.
The central challenge, therefore, lies in achieving fairness through \emph{directional improvement}---enhancing underperforming groups while preserving representational integrity and overall utility.

\textbf{(2) Incompatible fairness definitions.}
Fairness metrics are predominantly defined for discrete classification or detection \cite{zafar2017fairness,menon2018cost,berk2017convex}, which can not naturally extend to regression outputs. Different tasks operate under \emph{incompatible} fairness notions, and no unified mechanism exists to aggregate them into a single optimization objective;
\textbf{(3) Head-induced residual bias of MTL.}
Conventional MTL paradigms  (\textbf{Fig.~\ref{fig:framework}(d)}) constrain only the shared encoder while leaving task heads free to internalize residual bias. As each head independently reshapes gradients, group disparities re-emerge at the prediction level and remain uncontrolled across tasks, revealing a lack of fairness awareness and an inability to model head-induced geometry; and 
\textbf{(4) Fairness under partial supervision.}
Both fairness-aware and MTL methods implicitly assume \textit{full annotations} are provided across all tasks \cite{sener2018multi,lin2019pareto,mahapatra2020multi,yu2020gradient}, whereas real-world datasets contain non-overlapping labels. Under partial annotation, models lack consistent supervision and joint utility–fairness optimization becomes ill-posed.

To address the above challenges in developing fair MTL, we propose \textsc{FairMT}, the \textit{first} unified fairness-aware MTL framework that addresses all four challenges simultaneously.
For \textbf{Challenges (1)}, we introduces task-dependent \textit{asymmetric fairness disparity} metric that promote directional improvement for under-served groups rather than degrading dominant ones. Through an \textit{Asymmetric Heterogeneous Fairness Disparity Aggregation (AHFDA)} mechanism, these task-specific disparity are consolidated into a single, optimizable objective, which addresses both \textbf{Challenges (2)} and \textbf{Challenges (4)} by providing a principled aggregation rule that accommodates heterogeneous outputs and partial supervision.
\textbf{Challenge (3)} is solved by a \textit{head-aware optimization proxy}, which not only corrects head-induced geometric bias but also reduces the full-parameter constrained optimization to an efficient head-space surrogate, preserving fairness at a fraction of the computational cost. 
These components are integrated into a tractable primal--dual formulation, enabling joint optimization of utility and fairness under a unified theoretical framework.

\noindent
\textbf{The main contributions of this paper are fourfold:}
\begin{itemize}
    \item \textbf{\textsc{FairMT}:} The first unified fairness-aware MTL framework that seamlessly supports classification, detection, and regression, even under partial task annotation.
    \item \textbf{AHFDA:} An asymmetric heterogeneous fairness aggregation mechanism that yields directional fairness gains for under-served groups without inducing utility collapse.
    \item \textbf{Head-aware multi-objective optimization:} Eliminates residual head-level bias and provides a tractable descent geometry for fairness optimization.
    \item \textbf{SOTA performance.} Extensive experiments across heterogeneous benchmarks, spanning datasets from different modalities and supervision regimes, shows substantial fairness gains with superiority in utility, consistently outperformed existing leading fairness and MTL baselines.
\end{itemize}

\section{Preliminaries}

We consider a fairness-aware multi-task learning (MTL) setting with $T$ heterogeneous tasks, indexed by $[T]=\{1,\ldots,T\}$. 
The training corpus is compiled from heterogeneous datasets (e.g., valence-arousal regression and action-unit detection), resulting in inherently \textit{partial task supervision}—each sample provides labels for only a subset of tasks.
Each sample is written as $(X,G,\{Y_t\}_{t\in\Omega})$, where $X\in\mathcal{X}$ is the input, $G\in\mathcal{G}$ is a sensitive attribute, and $\Omega\subseteq[T]$ denotes the subset of tasks with observed labels.
The MTL model is parameterized by $\boldsymbol{\theta}=(\boldsymbol{\theta}_s,\{\boldsymbol{\theta}_t\}_{t=1}^T)$, where $\boldsymbol{\theta}_s$ denotes shared parameters and $\boldsymbol{\theta}_t$ denotes task-specific parameters. Each task $t$ predicts via
$f_{\boldsymbol{\theta}_s,\boldsymbol{\theta}_t}:\mathcal{X}\to\widehat{\mathcal{Y}}_t$
and incurs loss $\ell_{t,\boldsymbol{\theta}}(x,y_t)=M_t\,\ell_t(f_{\boldsymbol{\theta}_s,\boldsymbol{\theta}_t}(x),y_t)$, where $M_t=\mathbbm{1}[t\in\Omega]$ masks unobserved labels.

\noindent
\textbf{Conventional MTL.}
For each task $t$, assign a nonnegative weight $c_t$, with $\sum_{t=1}^T c_t>0$, the  MTL objective is
{
\setlength{\abovedisplayskip}{5pt}
\setlength{\belowdisplayskip}{8pt}
\begin{equation*}
\label{eq:mtl_objective}
\min_{\boldsymbol{\theta}_s,\{\boldsymbol{\theta}_t\}}
\sum_{t=1}^T c_t\,L_t(\boldsymbol{\theta}_s,\boldsymbol{\theta}_t),
\end{equation*}
where $L_t(\boldsymbol{\theta}_s,\boldsymbol{\theta}_t):=\mathbb{E}[\ell_{t,\boldsymbol{\theta}}(X,Y_t)]$ is task-specific risk.
}

\noindent
\textbf{Fairness in MTL.}
Classical fair learning imposes per-task constraints of the form
$\psi_t(\boldsymbol{\theta}_s,\boldsymbol{\theta}_t)\le\epsilon_t$,
where $\psi_t$ measures group disparities (e.g., equalized odds, demographic-parity relaxations). A direct MTL extension yields
\begin{equation*}
\label{eq:constrained_mtl}
\min_{\boldsymbol{\theta}_s,\{\boldsymbol{\theta}_t\}}
\sum_{t=1}^T c_t L_t(\boldsymbol{\theta}_s,\boldsymbol{\theta}_t)
\quad\text{s.t.}\quad
\sum_{t=1}^T \psi_t(\boldsymbol{\theta}_s,\boldsymbol{\theta}_t)\le\epsilon.
\end{equation*}

\noindent
\textbf{Differentiable Fairness Rates} \textsc{FairMT} requires a unified formulation of differentiable
fairness rates for both binary detection tasks and multi-class classification
tasks, ensuring consistent gradient signals across heterogeneous task types.  
We therefore define soft population-level rates and their mini-batch estimators for
each task type as follows.

\vspace{0.2em}
\textit{Binary Detection Task.} For each binary detection task \(t\in[T]\) with partial-annotation mask \(M_t=\mathbbm{1}[t\in\Omega]\), let
\(p_t(x)\in[0,1]\) denote the \emph{differentiable predicted-positive score}, typically parameterized as
\(p_t(x)=\sigma\!\big(f_{\boldsymbol{\theta}_s,\boldsymbol{\theta}_t}(x)\big)\)
with \(\sigma(\cdot)\) the logistic sigmoid.
For a sensitive group \(g\in\mathcal{G}\),
define the \emph{soft} true positive rate (TPR) and false positive rate (FPR) as the conditional expectations
$\mathrm{TPR}_{g,t}
= \mathbb{E}\!\left[p_t(X)\mid Y_t=1,\,G=g\right]$,
$\mathrm{FPR}_{g,t}
= \mathbb{E}\!\left[p_t(X)\mid Y_t=0,\,G=g\right]$.
Their mini-batch estimates on \(\mathcal{B}\) are naturally obtained via masked averages:
{
\footnotesize
\begin{equation*}
\begin{aligned}
\widehat{\mathrm{TPR}}_{g,t}
&=\frac{1}{|\mathcal{B}_{g,t}^{+}|}
 \sum_{i\in\mathcal{B}_{g,t}^{+}} p_t(X^{(i)}),\\
\widehat{\mathrm{FPR}}_{g,t}
&=\frac{1}{|\mathcal{B}_{g,t}^{-}|}
 \sum_{i\in\mathcal{B}_{g,t}^{-}} p_t(X^{(i)}),
\end{aligned}
\end{equation*}
}\noindent
where
\(\mathcal{B}_{g,t}^{+}=\{i\in\mathcal{B}:G^{(i)}=g,\,Y_t^{(i)}=1\}\)
and
\(\mathcal{B}_{g,t}^{-}=\{i\in\mathcal{B}:G^{(i)}=g,\,Y_t^{(i)}=0\}\)
denote the sets of positive and negative samples for group \(g\) and task \(t\), respectively.

\textit{Multi-Class Classification Task.} 
For task \(t\!\in[T]\) with partial-annotation mask \(M_t=\mathbbm{1}[t\in\Omega]\), let \(f_{\boldsymbol{\theta}_s,\boldsymbol{\theta}_t}(x)\in\mathbb{R}^{C_t}\) denote the class logits, and define temperature-scaled probabilities 
\(p_{t,k}(x)=\mathrm{softmax}(f_{\boldsymbol{\theta}_s,\boldsymbol{\theta}_t}(x))_k\),
where $k\in\{1,\dots,C_t\}$.  
For each class \(k\), we treat \(k\) as “positive’’ and all other classes as “negative’’,
yielding class-wise soft rates
{
\footnotesize
\begin{equation*}
\begin{aligned}
\widehat{\mathrm{TPR}}_{g,t,k}
&=\frac{1}{|\mathcal{B}_{g,t,k}^{+}|}
 \sum_{i\in\mathcal{B}_{g,t,k}^{+}} p_{t,k}(X^{(i)}),  \\
\widehat{\mathrm{FPR}}_{g,t,k}
&=\frac{1}{|\mathcal{B}_{g,t,k}^{-}|}
 \sum_{i\in\mathcal{B}_{g,t,k}^{-}} p_{t,k}(X^{(i)}),
\end{aligned}
\end{equation*}}\noindent
where $\mathcal{B}_{g,t,k}^{+}$ and 
$\mathcal{B}_{g,t,k}^{-}$ denote  positive and negative group.

\section{\textsc{FairMT} Framework}
\label{sec:FairMT}

In this section, we present \emph{\textsc{FairMT}} (\textbf{Fig.\ref{fig:framework}(a)}), a fairness-aware Pareto-guided primal–dual framework.

\subsection{Asymmetric Heterogeneous Fairness Disparity Aggregation (AHFDA)}
\label{sec:AHFDA}
For classification, detection, and regression tasks, we define task-specific asymmetric fairness disparities emphasizing the worst-performing group and integrate them through a tailored simplex-based optimization. 
Instead of averaging heterogeneous disparities, AHFCA concentrates constraint pressure on the most violated tasks, producing a unified fairness objective via projection-free simplex updates with sparse, vertex-structured iterates.

\subsubsection{Fair Binary Detection.}
For binary detection, group fairness is commonly measured by following metric:
\emph{(i) Demographic Parity} (DP) enforces equality of positive prediction rates across groups,
but ignores correctness and is unsuitable in settings where fairness must be conditioned on true labels.
\emph{(ii) Equal Opportunity} (EO) instead requires equal true positive rates (TPR),
while \emph{(iii) Equalized Odds} (EOD) additionally equalizes false positive rates (FPR),
In the sequel we adopt EOD as primary detection-fairness constraint (with EO recovered as a special case) and formalize its asymmetric variant below.

\vspace{0.2em}
\textbf{Asymmetric Equalized Odds (AEOD).}
Classical EOD \cite{garg2020fairness,tang2022attainability,zhong2024intrinsic} evaluate symmetric discrepancies such as 
\(|\mathrm{TPR}_{g,t}-\mathrm{TPR}_{g',t}|\) and \(|\mathrm{FPR}_{g,t}-\mathrm{FPR}_{g',t}|\), 
allowing fairness to be achieved by \emph{lowering} the performance of well-served groups rather than improving under-served ones. To enforce \emph{directional} improvement, each task \(t\) is anchored to a group-wise \textbf{reference }(advantated group), so only lagging groups are penalized while high-performing groups remain unaffected.
For true positives and false positives we define
{
\small
\begin{equation*}
\tau_t \;=\; \max_{g\in\mathcal{G}_t^{+}} \ \widehat{\mathrm{TPR}}_{g,t},
\qquad
\phi_t \;=\; \min_{g\in\mathcal{G}_t^{-}} \ \widehat{\mathrm{FPR}}_{g,t},
\end{equation*}
}\noindent
where \(\tau_t\) represents the best attainable TPR among all valid groups, 
and \(\phi_t\) is the lowest achievable FPR, \(\mathcal{G}_t^{+}=\{g:\ |\mathcal{B}_{g,t}^{+}|>0\}\) and \(\mathcal{G}_t^{-}=\{g:\ |\mathcal{B}_{g,t}^{-}|>0\}\) collect valid groups for positives and negatives. 
These references prevent performance collapse, ensuring fairness improvements are driven 
toward the current strongest group rather than suppressing it.
We then penalize \emph{only} lagging TPRs and \emph{only} exceeding FPRs via one-sided hinge gaps:
{
\footnotesize
\begin{equation*}
\begin{aligned}
\Delta^{\mathrm{TPR}}_{g,t}
=\big[\ \tau_t - \widehat{\mathrm{TPR}}_{g,t}\ \big]_+,
\qquad
\Delta^{\mathrm{FPR}}_{g,t}
=\big[\ \widehat{\mathrm{FPR}}_{g,t} - \phi_t\ \big]_+,
\end{aligned}
\end{equation*}
}\noindent
where \([z]_+=\max\{z,0\}\).
To aggregate group-wise gaps into a single per-task disparity, a worst-case reducer is used.
The AEOD $\psi^{\mathrm{AEOD}}_t$ for task \(t\) is defined as 
{
\footnotesize
\begin{equation}
M_t\,
\frac{1}{1+\beta }
\left(
\max_{g\in\mathcal{G}_t^{+}}\big(\Delta^{\mathrm{TPR}}_{g,t}\big)^{2}
+
\eta\,\max_{g\in\mathcal{G}_t^{-}}\big(\Delta^{\mathrm{FPR}}_{g,t}\big)^{2}
\right),
\end{equation}
}\noindent
where \(\beta \in\{0,1\}\) recovers EO (\(\beta =0\)) or EOD (\(\beta =1\)), and 
\(M_t=\mathbbm{1}[t\in\Omega]\) enforces partial annotation.

\subsubsection{Fair Multi-class Classification.}
Unlike the binary case, fairness in multi-class classification is evaluated in a one-vs-rest manner, treating each class independently against all others and reducing multi-class predictions to a sequence of aligned positive–negative decisions. The resulting disparities are typically measured using \textbf{Multi-EOD/EO}, extending equalized TPR/FPR criteria in the same spirit as fairness evaluation for detection tasks.
we adapt Multi-EOD to construct our asymmetric variant.

\vspace{0.2em}
\textbf{Class-wise Asymmetric Equalized Odds (CAEOD).}
As in the binary case, to enforce directional fairness, each class \(k\) is anchored to its best attainable reference level:
{\small
\begin{equation*}
\tau_{t,k}=\max_{g\in\mathcal{G}^{+}_{t,k}}\widehat{\mathrm{TPR}}_{g,t,k},
\qquad
\phi_{t,k}=\min_{g\in\mathcal{G}^{-}_{t,k}}\widehat{\mathrm{FPR}}_{g,t,k},
\end{equation*}}
and we penalize only lagging TPRs or exceeding FPRs through one-sided hinge gaps
{\small
\begin{equation*}
\begin{aligned}
\Delta^{\mathrm{TPR}}_{g,t,k}
&=[\,\tau_{t,k}-\widehat{\mathrm{TPR}}_{g,t,k}\,]_+,\\
\Delta^{\mathrm{FPR}}_{g,t,k}
&=[\,\widehat{\mathrm{FPR}}_{g,t,k}-\phi_{t,k}\,]_+.
\end{aligned}
\end{equation*}}
\noindent
For each class \(k\), group discrepancies are reduced via a worst-case operator and
squared to stabilize gradients:
{\small
\begin{equation}
\psi^{\mathrm{CAEOD}}_{t,k}
=
\frac{M_t}{1+\eta}
\left(
\max_{g}(\Delta^{\mathrm{TPR}}_{g,t,k})^2
+
\eta\,\max_{g}(\Delta^{\mathrm{FPR}}_{g,t,k})^2
\right),
\end{equation}}\noindent
where $k=1,\dots,C_t,$ \(\eta\in\{0,1\}\) recovers EO (\(\eta=0\)) or EOD (\(\eta=1\)).
Thus, a multi-class task produces a vector
\(
[\psi^{\mathrm{CAEOD}}_{t,1},\dots,\psi^{\mathrm{CAEOD}}_{t,C_t}],
\)
preserving per-class granularity and avoiding any implicit averaging across classes.

\subsubsection{Fair Regression.}
Group fairness in regression lacks a universally adopted metric. In practice, several criteria are commonly used:
\textbf{(i) KS} (KS Distributional Demographic Parity)\cite{chzhen2020fair,chzhen2020fair1}, requiring group-wise predictive distributions to match, i.e.,
\(\sup_{C\subset\mathbb{R}} \big| P(\hat{y}\!\in\!C\mid g) - P(\hat{y}\!\in\!C\mid g') \big| = 0\),
\textbf{(ii) CSP} (Conditional Statistical Parity)\cite{agarwal2019fair,wei2023mean},
Split by a threshold \(\tau\) and evaluate, for each stratum \(S\in\{Y\!\ge\!\tau,\,Y\!<\!\tau\}\),
\(\Delta_S = \max_g \mathbb{E}[\hat{Y}\mid G{=}g,S] - \min_g \mathbb{E}[\hat{Y}\mid G{=}g,S]\),
where smaller \(\Delta_S\) implies greater fairness.
\textbf{(iii) EP} (Error parity)\cite{agarwal2019fair}, enforcing similarity of group-wise expected errors \(e_g=\mathbb{E}[|y-\hat{y}|\mid g]\);
and \textbf{(iv) Binned EO/EOD}\cite{mehrabi2021survey}, continuous predictions \(\hat{y}\) are discretized into \(K\) bins, inducing a \(K\)-way one-vs-rest task with EO/EOD.

\textbf{Asymmetric Error Parity (AEP).} 
To unify all above heterogeneous metric within a single differentiable objective, we introduce \textbf{AEP}, which explicitly reduces group-wise error parity and thereby
jointly tightens multiple fairness criteria. In particular, shrinking residual gaps
between groups contracts marginal prediction gaps and improves distributional parity
(KS), aligns group-conditioned prediction trends and reduces $\mathbb{E}[\hat{Y}\mid G,S]$
discrepancies (CSP), and limits miscalibrated score bins that induce inter-group
$\mathrm{TPR}/\mathrm{FPR}$ differences (EO/EOD). 

Formally, for task \(t\in[T]\) with partial-annotation mask \(M_t=\mathbbm{1}[t\in\Omega]\), AEP impose a
\emph{one-sided}, fully differentiable penalty that only encourages under-performing groups
to catch up with the best-performing group. 
let \(g_t(x)=f_{\boldsymbol{\theta}_s,\boldsymbol{\theta}_t}(x)\in\mathbb{R}\) denote the predictor, and
\(\ell(\cdot,\cdot)\in\{\mathrm{MAE},\mathrm{MSE}\}\) be the regression loss.
On a mini-batch \(\mathcal{B}\), define sample sets $\mathcal{B}_{g,t}=\{\,i\in\mathcal{B}:G^{(i)}=g\,\}$, and empirical group errors
$\widehat{\mathrm{Err}}_{g,t}
=
\frac{1}{|\mathcal{B}_{g,t}|}
\sum_{i\in\mathcal{B}_{g,t}}
\ell\!\left(g_t(X^{(i)}),Y_t^{(i)}\right)$,
and collect valid groups as \(\mathcal{G}_t=\{g:\ |\mathcal{B}_{g,t}|>0\}\).
The best achievable performance on task \(t\) is used as a reference, and only groups exceeding this are penalized:
\begin{equation*}
\widehat{\Delta}_{g,t}
=\big[\widehat{\mathrm{Err}}_{g,t}-\hat e_t^{\mathrm{ref}}\big]_+,
\qquad
\hat e_t^{\mathrm{ref}}=\min_{g\in\mathcal{G}_t}\widehat{\mathrm{Err}}_{g,t}.
\end{equation*}
Larger deviations indicate greater disparity, and a worst-case aggregation drives uniform improvement across lagging groups.
The resulting constraint is
\begin{equation}
\psi^{\mathrm{AEP}}_t
=
M_t\,\max_{g\in\mathcal{G}_t}\big(\widehat{\Delta}_{g,t}\big)^2,
\end{equation}

\subsubsection{Fairness Disparity Aggregation}
\label{subsec:AHFDA}
In \textsc{FairMT}, each task \(t\!\in\![T]\) produces a nonnegative violation score \(\psi_t\) from the asymmetric metric defined above. 
Simply summing or averaging \(\{\psi_t\}\) is misaligned with multi-task fairness: small disparities can mask severe ones, effort may be wasted on already-fair tasks, and heterogeneity in task difficulty is ignored. 
An effective aggregator should therefore (i) concentrate pressure on most violated tasks, (ii) be directional, avoiding penalties on fair tasks, (iii) adaptively allocates optimal pressure across tasks.

\textbf{AHFDA Objective Formulation.}
To aggregate per-task violations into a single fairness objective, we assign each \(\psi_t\) a weight \(w_t\) with \(\sum_t w_t=1\), ensuring a normalized and interpretable allocation of corrective pressure across tasks. Let \(\boldsymbol{\Psi}=(\psi_1,\ldots,\psi_T)\in\mathbb{R}^T_{\ge 0}\) denote task-level violations and \(\boldsymbol{w}\in\Delta_T=\{\boldsymbol{w}\!\ge\!0,\ \mathbf{1}^\top\boldsymbol{w}=1\}\) be simplex weights. We define the \emph{directional, heterogeneity-aware} aggregator
\begin{equation}
\label{eq:AHFDA-max}
\Phi(\boldsymbol{\Psi})
=\max_{\boldsymbol{w}\in\Delta_T}
\ \boldsymbol{w}^{\top}\boldsymbol{\Psi}
-\frac{\mu}{2}\,\lVert\boldsymbol{w}\rVert_2^2,
\qquad \mu>0.
\end{equation}
The linear term concentrates mass on high-violation tasks, enforcing a worst-case bias; the quadratic term prevents collapse to a single task, yielding stable and smooth weights. $\mu$ controls the \emph{aggressiveness} of weight reallocation: small $\mu$ concentrates on high-$\psi$ tasks, sharpening worst-case correction, whereas larger $\mu$ distributes pressure more smoothly. 

\vspace{0.2em}
\textbf{Projection View and Closed-Form Solution.}
To obtain an analytically tractable characterization of the optimal simplex weight, we rewrite the maximization in Eq.~\eqref{eq:AHFDA-max} as the equivalent strongly convex minimization
{\small
\begin{equation}
\label{eq:AHFDA-min}
\boldsymbol{w}^\star
=\frac{\mu}{2}\|\boldsymbol{w}\|_2^2 - \boldsymbol{\Psi}^\top\boldsymbol{w},
\qquad \mu>0,
\end{equation}}\noindent
which yields interpolation form
$\frac{\mu}{2}\Bigl\|\boldsymbol{w}-\tfrac{1}{\mu}\boldsymbol{\Psi}\Bigr\|_2^2
+\text{const}$,
Hence, the minimizer admits a \textbf{closed-form} \emph{ geometric characterization} as the Euclidean projection of $\tfrac{1}{\mu}\boldsymbol{\Psi}$ onto the probability simplex:
$\boldsymbol{w}^\star
= \operatorname{Proj}_{\Delta_T}\!\Bigl(\tfrac{1}{\mu}\boldsymbol{\Psi}\Bigr)$,
there exists a scalar Lagrange multiplier $\tau\in\mathbb{R}$ such that 
\begin{equation}
\label{eq:water-filling}
w_t^\star
=\max\!\bigl\{\,0,\ \tfrac{\psi_t}{\mu}-\tau\,\bigr\},
\qquad
\sum_{t=1}^T w_t^\star = 1.
\end{equation}\noindent
This closed-form expression is the well-known \emph{water-filling} structure: coordinates falling below the learned threshold $\tau$ are truncated to $0$, while the remaining coordinates are linearly shifted to satisfy the simplex constraint. Consequently, AHFDA is intrinsically \emph{directional}: tasks with negligible or vanishing violations (i.e., $\psi_t\!\approx\!0$) are assigned zero weight, concentrating the fairness signal on the most severe violators (see \textbf{\textit{Appendix~\textcolor{red}{A}}} for a complete derivation).  

\vspace{0.2em}
\textbf{Projection-Free Aggregation.}
While Eq.\eqref{eq:water-filling} gives an exact solution, repeated projections can be computationally heavy and may destroy the sparsity of weight iterates within a training loop. We instead adopt a \emph{projection-free} conditional gradient (Frank--Wolfe, FW) solver tailored to Eq.\eqref{eq:AHFDA-min}. Writing Eq.\eqref{eq:AHFDA-min} as a quadratic program
\begin{equation}
\label{eq:qp}
\min_{\boldsymbol{w}\in\Delta_T}
\frac12\,\boldsymbol{w}^\top(\mu I)\boldsymbol{w}-\boldsymbol{\Psi}^\top\boldsymbol{w},
\quad
\nabla f(\boldsymbol{w})=\mu\boldsymbol{w}-\boldsymbol{\Psi},
\end{equation}
it requires only a \textit{linear minimization oracle (LMO)} over the simplex, keeping iterates as sparse convex combinations of vertices and yielding interpretable active sets of unfair tasks.

The full procedure of projection-free AHFDA is given in \textbf{Algorithm~\ref{alg:fw_fairness}}.
At iteration $\ell$, AHFDA constructs a feasible search direction by linearizing the objective
around the current iterate $\boldsymbol{w}^{(\ell)}\!\in\!\Delta_T$:
$\min_{\boldsymbol{s}\in\Delta_T}
\ \big\langle\nabla f(\boldsymbol{w}^{(\ell)}),\, \boldsymbol{s}\big\rangle$,
where $\langle\cdot,\cdot\rangle$ denotes the Euclidean inner product and
$\boldsymbol{s}$ is a feasible point on the simplex (i.e., a candidate descent direction).
Since $\Delta_T$ is the convex hull of $\{e_1,\ldots,e_T\}$, any linear objective over the simplex
attains its minimum at one of these vertices. Consequently, the LMO returns
\begin{equation}
\label{eq:lmo-short}
\boldsymbol{s}^{(\ell)} = e_{j_\ell},
\qquad
j_\ell \in \arg\max_{t\in\{1,\ldots,T\}} \bigl(\psi_t - \mu w^{(\ell)}_t\bigr),
\end{equation}
where $e_{j_\ell}$ is the simplex vertex corresponding to task $j_\ell$.
The iterate is then updated by a convex combination, with a step size $\gamma_\ell\!\in\![0,1]$ to maintain feasibility:
\begin{equation}
\label{eq:fw-update-short}
\boldsymbol{w}^{(\ell+1)}
= (1-\gamma_\ell)\,\boldsymbol{w}^{(\ell)}
+ \gamma_\ell\,\boldsymbol{s}^{(\ell)} \in \Delta_T.
\end{equation}
For quadratic objective Eq.\,\eqref{eq:AHFDA-min}, step size admits a closed form.
Let $\boldsymbol{d}^{(\ell)}=\boldsymbol{s}^{(\ell)}-\boldsymbol{w}^{(\ell)}$. Exact line search gives
\begin{equation}
\label{eq:fw-stepsize-short}
\gamma_\ell
= \Pi_{[0,1]}\!\left(
-\frac{\langle\,\mu\boldsymbol{w}^{(\ell)}-\boldsymbol{\psi},\ \boldsymbol{d}^{(\ell)}\,\rangle}
       {\mu\,\|\boldsymbol{d}^{(\ell)}\|_2^2}
\right),
\end{equation}
where $\Pi_{[0,1]}$ denotes projection onto $[0,1]$. Thus, each AHFDA iteration consists of a gradient
evaluation, one coordinate maximization, and a scalar step-size update—an $O(T)$ procedure that
preserves sparsity and interpretability of the fairness weights (Detailed derivations and
convergence guarantees are provided in \textbf{\textit{Appendix~\textcolor{red}{B}}}).


\begin{algorithm}[t]
\caption{Asymmetric Heterogeneous Fairness Disparity Aggregation (AHFDA)}
\label{alg:fw_fairness}
\begin{algorithmic}[1]
\REQUIRE Violations $\boldsymbol{\Psi}\in\mathbb{R}^d_{\ge0}$, $\mu>0$, tolerance $\varepsilon$
\ENSURE $\boldsymbol{w}^\star\in\Delta_T$
\STATE Initialize $\boldsymbol{w}^{(0)}\in\Delta_T$ (e.g., uniform), set $t=0$
\REPEAT
    \STATE $\nabla f(\boldsymbol{w}^{(t)})=\mu\boldsymbol{w}^{(t)}-\boldsymbol{\Psi}$
    \STATE $j_t \in \arg\max_{k\in[d]} (\psi_k-\mu w_k^{(t)})$,\quad $\boldsymbol{s}^{(t)}=e_{j_t}$
    \STATE $\boldsymbol{d}^{(t)}=\boldsymbol{s}^{(t)}-\boldsymbol{w}^{(t)}$
    \STATE \textbf{Exact line-search:}     {\small
    \[
    \smash{\gamma_t
    = \Pi_{[0,1]}\!\left(
    -\frac{
        \langle \mu\boldsymbol{w}^{(t)}-\boldsymbol{\Psi},\; \boldsymbol{d}^{(t)} \rangle
    }{
        \mu\,\|\boldsymbol{d}^{(t)}\|_2^2
    }\right)}
    \]}
    \STATE $\smash{\boldsymbol{w}^{(t+1)} = \boldsymbol{w}^{(t)} + \gamma_t\,\boldsymbol{d}^{(t)} \in\Delta_T}$
    \STATE $g_t=\langle\nabla f(\boldsymbol{w}^{(t)}),\, \boldsymbol{w}^{(t)}-\boldsymbol{s}^{(t)}\rangle$
    \STATE $t\leftarrow t+1$
\UNTIL{$g_{t-1}\le\varepsilon$}
\RETURN $\boldsymbol{w}^\star=\boldsymbol{w}^{(t)}$
\end{algorithmic}
\end{algorithm}

\subsection{\textsc{\textbf{FairMT}} Framework}
\label{sec:primal-dual}
In conventional MTL without fairness, the training objective is cast as a multi-objective optimization (MOO) balancing task risks~\cite{desideri2012multiple,sener2018multi}. Once the aggregated fairness disparity $\Phi(\boldsymbol{\theta})$ from AHFDA in Eq.\,\eqref{eq:AHFDA-max} is imposed as a constraint $\Phi(\boldsymbol{\theta})\le\epsilon$, the problem becomes a \emph{constrained} MOO:
{\small
\begin{equation*}
\min_{\boldsymbol{\theta}_s,\{\boldsymbol{\theta}_t\}}
\;\big(L_1(\boldsymbol{\theta}_s,\boldsymbol{\theta}_1),\ldots,L_T(\boldsymbol{\theta}_s,\boldsymbol{\theta}_T)\big),
\quad 
\text{s.t. } \Phi(\boldsymbol{\theta}) \le \epsilon.
\end{equation*}}\noindent

To address this, we introduce \textbf{\textsc{FairMT}}, a \emph{Fairness-Aware Pareto-Guided Primal--Dual}  framework, 
which introduces \emph{\textbf{primal}} scalarization weights to balance the tasks and a \emph{\textbf{dual}} multiplier to enforce the fairness constraint, relaxing the constrained formulation into an unconstrained one. This yields a saddle-point reformulation whose stationary solutions coincide with Pareto-stationary points.

Specifically, we seek to update the multi-task utility along a common descent direction that simultaneously improves all tasks. In line with AHFDA, we introduce the standard simplex
$\Delta_T := \bigl\{\boldsymbol{\alpha}\in\mathbb{R}_{\ge 0}^T \,\big|\, \sum_{t=1}^T \alpha_t = 1 \bigr\}$,
and let $\boldsymbol{\alpha}\in\Delta_T$ denote the primal task weights that govern navigation along the Pareto frontier. The simplex constraint enforces nonnegativity and unit-sum normalization, ensuring that $\boldsymbol{\alpha}$ encodes valid convex combinations of task gradients.

Let the model parameters be $\boldsymbol{\theta} := \bigl(\boldsymbol{\theta}_s,\{\boldsymbol{\theta}_t\}_{t=1}^T\bigr)$, aggregated fairness disparity $\Phi(\boldsymbol{\theta})$, and tolerance $\epsilon \ge 0$ for admissible violations. The fairness constraint is therefore written as
$h(\boldsymbol{\theta}) \triangleq \Phi(\boldsymbol{\theta}) - \epsilon \le 0$.
We form the Lagrangian
\begin{equation}
\mathcal{L}(\boldsymbol{\theta},\boldsymbol{\alpha},\eta)
=
\underbrace{\sum_{t=1}^T \alpha_t\,L_t(\boldsymbol{\theta}_s,\boldsymbol{\theta}_t)}_{\substack{\text{primal (task risks)}}}
\;+\;
\underbrace{\eta\,h(\boldsymbol{\theta})}_{\substack{\text{dual (fairness)}}}.
\label{eq:Lagrangian}
\end{equation}
Here, $\eta$ is the dual multiplier associated with $h(\boldsymbol{\theta})$, adaptively regulating the influence of the fairness constraint and yielding a saddle-point formulation that harmonizes multi-task learning with constraint satisfaction.
Consequently, the \textsc{FairMT} objective can be written in min--max form:
\begin{equation}
\min_{\boldsymbol{\theta}_s,\,\{\boldsymbol{\theta}_t\}_{t=1}^T,\,\boldsymbol{\alpha}\in\Delta_T}
\;\max_{\eta\ge 0}\;
\mathcal{L}\bigl(\boldsymbol{\theta}_s,\{\boldsymbol{\theta}_t\},\boldsymbol{\alpha},\eta\bigr).
\label{eq:minmax-new}
\end{equation}

\noindent
\textbf{KKT conditions and Pareto stationarity.}
An optimal saddle point $(\boldsymbol{\theta}_s^\star,\{\boldsymbol{\theta}_t^\star\}_{t=1}^T,\boldsymbol{\alpha}^\star,\eta^\star)$ satisfies:
\begin{enumerate}[label=(\roman*), leftmargin=*, labelsep=0.5em, align=left, itemsep=2pt]
\item \emph{Primal feasibility (utility):}
$h(\boldsymbol{\theta}^\star)\le 0$ and $\boldsymbol{\alpha}^\star\in\Delta_T$, i.e., $\Phi(\boldsymbol{\theta}^\star)\le\epsilon$, $\alpha_t^\star\ge 0$, and $\sum_{t=1}^T \alpha_t^\star=1$.
\item \emph{Dual feasibility (fairness):}
$\eta^\star\ge 0$. 
The dual multiplier assigns a nonnegative penalty to fairness violations, as required for inequality constraints.
\item \emph{Complementary slackness (fairness):} $\eta^\star\,h(\boldsymbol{\theta}^\star)=0$. 
If the constraint is active ($\Phi(\boldsymbol{\theta}^\star)=\epsilon$), then $\eta^\star\ge 0$; otherwise ($\Phi(\boldsymbol{\theta}^\star)<\epsilon$), multiplier vanishes ($\eta^\star=0$).
\item \emph{Stationarity (w.r.t.\ model parameters):} 
At optimality, first-order balance holds between aggregated task risks and fairness term. 
In \emph{shared} parameter block:
{\small
\setlength{\abovedisplayskip}{2pt}
\setlength{\belowdisplayskip}{7pt}
\begin{equation}
\label{eq:shared}
\nabla_{\boldsymbol{\theta}_s}\!\left(\sum_{t=1}^T \alpha_t^\star L_t(\boldsymbol{\theta}_s^\star,\boldsymbol{\theta}_t^\star)\right)
+\eta^\star\,\nabla_{\boldsymbol{\theta}_s}\Phi(\boldsymbol{\theta}^\star)=\boldsymbol{0},
\end{equation}}
and in each \emph{task-specific} block, for $\forall\,t\in\{1,\dots,T\}$,
\begin{equation}
\alpha_t^\star\,\nabla_{\boldsymbol{\theta}_t} L_t(\boldsymbol{\theta}_s^\star,\boldsymbol{\theta}_t^\star)
+\eta^\star\,\nabla_{\boldsymbol{\theta}_t}\Phi(\boldsymbol{\theta}^\star)=\boldsymbol{0}.
\end{equation}
\end{enumerate}

\noindent
Under these conditions, $(\boldsymbol{\theta}^\star,\boldsymbol{\alpha}^\star,\eta^\star)$ constitutes a \emph{fairness-constrained Pareto-stationary point}: 
the task risks are Pareto-balanced by $\boldsymbol{\alpha}^\star$ while feasibility of the fairness constraint is enforced via $\eta^\star$. 

\subsection{\textsc{\textbf{FairMT}} Optimization}
\label{sec:optimization}
Since \textsc{FairMT} jointly optimizes task risks under a global fairness constraint, exact stationary is generally intractable. We adopt an iterative primal–dual optimization scheme.

\textbf{STEP 1: Primal Update of Task Weights.}
The task weights $\boldsymbol{\alpha}$ are explicit optimization variables in the KKT Eq.~\eqref{eq:shared}. They control how task gradients mix in the shared encoder and thus shape the descent geometry. 
However, the stationary point may not correspond to a valid simplex point $\boldsymbol{\alpha}\in\Delta_T$. We therefore approximate it by the \textit{least-squares projection} of fairness-adjusted direction onto the \textit{convex hull} of task gradients. Geometrically, with $\boldsymbol{\theta}$ fixed, the set $\{\nabla_{\boldsymbol{\theta}_s}L_t\}_{t=1}^T$ spans all feasible shared-parameter descent directions, and selecting best $\boldsymbol{\alpha}$ amounts to finding the convex combination closest to $-\eta\,\nabla_{\boldsymbol{\theta}_s}\Phi$, which is classical minimum-norm point problem~\citep{wolfe1976finding,sekitani1993recursive}. 
Thus, at iteration $r$ we compute $\boldsymbol{\alpha}^{(r)}$ as the minimizer of
{\small
\setlength{\abovedisplayskip}{5pt}
\setlength{\belowdisplayskip}{7pt}
\begin{equation}
\label{eq:w-ls-residual}
\min_{\boldsymbol{\alpha}\in\Delta_T}
\tfrac12\Big\|
\sum_{t=1}^T \alpha_t\,\nabla_{\boldsymbol{\theta}_s}L_t(\boldsymbol{\theta}^{(r)})
+\eta^{(r)}\,\nabla_{\boldsymbol{\theta}_s}\Phi(\boldsymbol{\theta}^{(r)})
\Big\|_2^2.
\end{equation}}\noindent
When $\eta^{(r)}{=}0$, it reduces exactly to the MGDA update~\cite{sener2018multi}.
When $T$ is large, solving the high-dimensional minimum-norm point exactly becomes costly and numerically fragile. We therefore introduce the \textbf{Head-Aware Proxy} in \textbf{Section.~\ref{sec:head-aware}}, which recasts residual into convex quadratic program (QP) of Eq.\,\eqref{eq:headaware_qp} and yields a stable, efficient update.

\vspace{0.2em}
\textbf{STEP 2: Primal Update of Model Parameters.}
Given the current task weights $\boldsymbol{\alpha}^{(r)}$, we update the shared parameters $\boldsymbol{\theta}_s$ and the task-specific parameters 
$\{\boldsymbol{\theta}_t\}_{t=1}^T$ by a gradient step on the Lagrangian. 
With learning rate $\rho>0$,
{\small
\setlength{\abovedisplayskip}{5pt}
\setlength{\belowdisplayskip}{5pt}
\begin{align}
\boldsymbol{\theta}_s 
&\leftarrow \boldsymbol{\theta}_s
- \rho\Big(\sum_{t=1}^T \alpha_t^{(r)}\,\nabla_{\boldsymbol{\theta}_s}L_t(\boldsymbol{\theta})
+ \eta^{(r)}\,\nabla_{\boldsymbol{\theta}_s}\Phi(\boldsymbol{\theta})\Big),\\
\boldsymbol{\theta}_t 
&\leftarrow \boldsymbol{\theta}_t
- \rho\Big(\nabla_{\boldsymbol{\theta}_t}L_t(\boldsymbol{\theta})
+ \eta^{(r)}\,\nabla_{\boldsymbol{\theta}_t}\Phi(\boldsymbol{\theta})\Big),
\quad \forall t\in[T].
\end{align}
}\noindent
This corresponds to a primal descent step under the fairness-adjusted gradient field.

\vspace{0.2em}
\textbf{STEP 3: Dual Update of the Fairness Multiplier.}
The dual variable $\eta$ is updated by a projected gradient-ascent step on the fairness constraint. 
With step size $\rho_\eta>0$,
{
\setlength{\abovedisplayskip}{2pt}
\setlength{\belowdisplayskip}{7pt}
\begin{equation}
\label{eq:dual_update}
\eta \;\leftarrow\; \big[\eta + \rho_\eta\big(\Phi_\varepsilon(\boldsymbol{\theta})-\epsilon\big)\big]_+,
\end{equation}}
where $[\cdot]_+$ projects onto $\mathbb{R}_{\ge 0}$. 
This update preserves dual feasibility and implicitly enforces complementary slackness: 
when $\Phi_\varepsilon(\boldsymbol{\theta})<\epsilon$, $\eta$ is driven toward $0$; 
when the constraint is violated, $\eta$ increases, strengthening the fairness term in the primal updates. 
Thus, the dual dynamics automatically regulate the degree of fairness enforcement.

\subsection{Head-Aware Optimization Proxy.}
\label{sec:head-aware}
In \textbf{STEP 1}, when $T$ is large, obtaining a fairness-aware multi-task descent direction requires full-model gradients for all tasks and the fairness functional, which is prohibitive in high-dimensional spaces. Prior methods introduce surrogates such as \textsc{MGDA-UB}~\cite{sener2018multi} construct a quadratic proxy by linearizing task losses in the shared representation, but implicitly assume that task heads are \textbf{i) geometrically neutral} and \textbf{ii) ignore the fairness constraint}. In practice, classifier heads apply heterogeneous linear transformations to shared features, distorting the effective descent geometry and potentially biasing updates. 
To address both issues, we introduce the \textbf{\textit{Head-Aware Optimization Proxy}}, which explicitly incorporates head-induced  geometry of task heads and their data-dependent utility-fairness sensitivities, and recasts residual into convex quadratic program of Eq.\,\eqref{eq:headaware_qp}. 

Let \(M_t=\mathbbm{1}[t\in\Omega]\) indicate which samples in the mini-batch contribute to task \(t\),
and define \(\mathcal{B}_t=\{i\in\mathcal{B}:M_t^{(i)}=1\}\).
For each task \(t\), let \(z_t(X^{(i)})\) denote the logit from
\(f_{\boldsymbol{\theta}_s,\boldsymbol{\theta}_t}\).
We compute the average logit-level sensitivities of the task losses and the fairness functional as
{\small
\begin{equation}
u_t=\frac{1}{|\mathcal{B}_t|}\sum_{i\in\mathcal{B}_t}
\frac{\partial L_t(\boldsymbol{\theta})}{\partial z_t(X^{(i)})},
\quad
(u_\Phi)_t=\frac{1}{|\mathcal{B}_t|}\sum_{i\in\mathcal{B}_t}
\frac{\partial \Phi(\boldsymbol{\theta})}{\partial z_t(X^{(i)})}.
\end{equation}}\noindent
Collecting \(\{u_t\}_{t=1}^T\), \(\{(u_\Phi)_t\}_{t=1}^T\) yields vectors
\(\boldsymbol{u}\), \(\boldsymbol{u}_\Phi\).

Let \(W\in\mathbb{R}^{T\times D}\) stack the final linear head weights of the \(T\) task heads.
The matrix \(W W^\top \in \mathbb{R}^{T\times T}\) is a Gram matrix: its \((i,j)\)-entry
\((W W^\top)_{ij}=W_i W_j^\top\) equals the inner product between the head directions of tasks \(i\) and \(j\) in the shared representation space.  
Large off-diagonal values thus indicate that two heads produce similar linear responses to shared features, while small values indicate weak alignment.
To incorporate data-dependent sensitivity, define \(\mathrm{Diag}(\boldsymbol{u})\) as the diagonal matrix with \((t,t)\)-entry \(u_t\).
We introduce the \emph{head-aware Gram matrix}
{
\begin{equation}
K \;:=\; \mathrm{Diag}(\boldsymbol{u})\,\big(W W^\top\big)\,\mathrm{Diag}(\boldsymbol{u}).
\end{equation}}
The diagonal reweighting preserves the geometry of \(W W^\top\) while amplifying tasks with higher logit sensitivity.  

To inject \textit{fairness} in the same geometry, we define
{
\begin{equation}
\boldsymbol{b}
\;:=\;
\eta\,K\Big(\boldsymbol{u}_\Phi\oslash(\boldsymbol{u}+\varepsilon\boldsymbol{1})\Big)\odot\boldsymbol{u},
\end{equation}}
where \(\eta\ge 0\) is the dual multiplier in Eq.\eqref{eq:dual_update}, \(\oslash\) denotes elementwise division, \(\odot\) the Hadamard product, and \(\varepsilon>0\) avoids division by zero.
Here \(\boldsymbol{u}_\Phi\in\mathbb{R}^T\) is the fairness analogue of \(\boldsymbol{u}\), whose \(t\)-th entry measures the average sensitivity of \(\Phi\) to the logits of task \(t\).
Consequently, utility and fairness live in a common inner-product space, preventing misaligned gradient updates.
To solve Eq.~\eqref{eq:w-ls-residual}, we project the shared gradients onto the head space.
Using the approximations
\(\nabla_{\boldsymbol{\theta}_s}L_t \approx u_t\,W_t\) and
\(\nabla_{\boldsymbol{\theta}_s}\Phi \approx (u_\Phi)_t\,W_t\),
the residual becomes a quadratic in \(\boldsymbol{\alpha}\) involving
\((K,\boldsymbol{b})\).  
This gives head-aware convex QP
{
\begin{equation}
\label{eq:headaware_qp}
\min_{\boldsymbol{\alpha}\in\Delta_T}\;
\tfrac12\,\boldsymbol{\alpha}^\top K\,\boldsymbol{\alpha}
+ \boldsymbol{b}^\top \boldsymbol{\alpha},
\end{equation}}
whose solution yields a head-aware approximation of the fairness-adjusted descent direction.
The problem involves only \(T\) variables and admits projection-free Frank--Wolfe updates at \(O(T^2)\) per iteration, negligible relative to backpropagation.  
When \(WW^\top\propto I_T\) and \(\boldsymbol{u}\propto\boldsymbol{1}\), the proxy reduces to \textsc{MGDA-UB}, recovering shared-only surrogate (Detailed derivations are provided in \textbf{\textit{Appendix~\textcolor{red}{C}}}).

\section{Experiments}\label{sec:exp}

\begin{table}[t]
\caption{
Comparison of \textsc{FairMT} with multi-task (\textcolor{red}{$^\dagger$}), fairness-only (\textcolor{red}{$^*$}), and hybrid fair-MTL (\textcolor{red}{$^{\dagger*}$}) baselines. 
Utility metrics (Acc, F1, CCC) are denoted with \textcolor{red}{$\uparrow$} (higher is better), while fairness metrics (EOD, EO, KS, CSP) are denoted with \textcolor{blue}{$\downarrow$} (lower is fairer).
}
\label{tab:sota}
\centering
\captionsetup{justification=centering}
\vspace{-0.5em}
\noindent\textbf{\footnotesize (a) Attribute Detection (in \%)}

\resizebox{\linewidth}{!}{
\setlength{\tabcolsep}{16pt}
\renewcommand{\arraystretch}{0.3}
\begin{tabular}{l|ccc|ccc}
\toprule
\multicolumn{7}{c}{\textbf{CelebA (Binary Detection)}} \\ 
\midrule
\multirow{2}{*}{Method} &\multirow{2}{*}{Acc \textcolor{red}{$\uparrow$}} & \multicolumn{2}{c|}{Age} & \multirow{2}{*}{Acc \textcolor{red}{$\uparrow$}} & \multicolumn{2}{c}{Gender} \\
\cmidrule(lr){3-4} \cmidrule(lr){6-7}
 &  & EOD \textcolor{blue}{\textdownarrow} & EO \textcolor{blue}{\textdownarrow} & & EOD \textcolor{blue}{\textdownarrow} & EO \textcolor{blue}{\textdownarrow} \\
\midrule
MGDA\textcolor{red}{$^\dagger$}       & 87.7 & 11.4 & 10.4 & 87.7 & 22.5 & 22.3 \\
GradNorm\textcolor{red}{$^\dagger$}   & 84.8 & 9.4  & 8.4  & 84.8 & 19.4 & 18.3 \\
\midrule
LNL\textcolor{red}{$^*$}        & 80.8 & 7.8  & 7.5  & 79.7 & 13.6 & 10.6 \\
UFaTE\textcolor{red}{$^*$}      & 76.4 & 6.8  & 5.5  & 75.4 & 12.2 & 10.2 \\
\midrule
MGDA-Mean\textcolor{red}{$^{\dagger*}$}  & 76.9 & 7.7  & 6.3  & 74.6 & 17.8 & 17.2 \\
MGDA-UFaTE\textcolor{red}{$^{\dagger*}$} & 78.6 & 6.8  & 4.4  & 77.8 & 12.6 & 9.6  \\
\midrule
\rowcolor{gray!20} 
\textbf{FairMT} & \textbf{88.4} & \textbf{3.4} & \textbf{2.0} & \textbf{88.1} & \textbf{6.4} & \textbf{3.9} \\
\bottomrule
\end{tabular}}

\vspace{1em}
\noindent\textbf{\footnotesize (b) Affective Analysis ( in \%)}

\resizebox{\linewidth}{!}{
\setlength{\tabcolsep}{7pt}
\renewcommand{\arraystretch}{0.4}
\begin{tabular}{l|ccc|ccc|ccc}
\toprule
\multicolumn{10}{c}{\textbf{AffectNet (Classification)}} \\
\midrule
\multirow{2}{*}{Method} & \multirow{2}{*}{Acc \textcolor{red}{$\uparrow$}} & \multicolumn{2}{c|}{Age} & \multirow{2}{*}{Acc \textcolor{red}{$\uparrow$}} &\multicolumn{2}{c|}{Gender} & \multirow{2}{*}{Acc \textcolor{red}{$\uparrow$}} &\multicolumn{2}{c}{Race} \\
\cmidrule(lr){3-4} \cmidrule(lr){6-7} \cmidrule(lr){9-10}
 &  & EOD \textcolor{blue}{\textdownarrow} & EO \textcolor{blue}{\textdownarrow} &  & EOD \textcolor{blue}{\textdownarrow} & EO \textcolor{blue}{\textdownarrow} &  & EOD \textcolor{blue}{\textdownarrow} & EO \textcolor{blue}{\textdownarrow} \\
\midrule
MGDA\textcolor{red}{$^\dagger$}       & 59.5 & 23.4 & 23.4 & \textbf{59.5} & 8.6  & 8.6  & \textbf{59.5} & 13.9 & 13.9 \\
GradNorm\textcolor{red}{$^\dagger$}   & 57.5 & 20.3 & 19.6 & 57.5 & 9.1  & 9.1  & 57.5 & 14.9 & 14.8 \\
\midrule
LNL\textcolor{red}{$^*$}        & 53.6 & 15.1 & 15.9 & 55.4 & 7.1  & 7.0  & 53.6 & 10.9 & 9.9  \\
UFaTE\textcolor{red}{$^*$}      & 51.1 & 12.8 & 12.4 & 53.4 & 6.0  & 6.0  & 54.0 & 8.7  & 8.5  \\
\midrule
MGDA-Mean\textcolor{red}{$^{\dagger*}$}  & 51.4 & 19.9 & 18.4 & 56.7 & 7.2  & 6.0  & 55.8 & 8.4  & 7.4  \\
MGDA-UFaTE\textcolor{red}{$^{\dagger*}$} & 55.6 & 15.0 & 13.2 & 56.4 & 6.7  & 6.8  & 56.2 & 7.4  & 7.4  \\
\midrule
\rowcolor{gray!20} 
\textbf{FairMT} & \textbf{60.6} & \textbf{9.9} & \textbf{8.3} & \textbf{59.5} & \textbf{4.0} & \textbf{3.9} & \underline{59.4} & \textbf{5.6} & \textbf{4.8} \\
\bottomrule
\end{tabular}}

\resizebox{\linewidth}{!}{
\setlength{\tabcolsep}{7pt}
\renewcommand{\arraystretch}{0.4}
\begin{tabular}{l|ccc|ccc|ccc}
\toprule
\multicolumn{10}{c}{\textbf{EmotioNet (Binary Detection)}} \\
\midrule
\multirow{2}{*}{Method} & \multirow{2}{*}{Acc \textcolor{red}{$\uparrow$}} & \multicolumn{2}{c|}{Age} & \multirow{2}{*}{Acc \textcolor{red}{$\uparrow$}} &\multicolumn{2}{c|}{Gender} & \multirow{2}{*}{Acc \textcolor{red}{$\uparrow$}} &\multicolumn{2}{c}{Race} \\
\cmidrule(lr){3-4} \cmidrule(lr){6-7} \cmidrule(lr){9-10}
 &  & EOD \textcolor{blue}{\textdownarrow} & EO \textcolor{blue}{\textdownarrow} &  & EOD \textcolor{blue}{\textdownarrow} & EO \textcolor{blue}{\textdownarrow} &  & EOD \textcolor{blue}{\textdownarrow} & EO \textcolor{blue}{\textdownarrow} \\
\midrule
MGDA\textcolor{red}{$^\dagger$}       & 81.3 & 19.1 & 19.4 & 81.3 & 6.5  & 5.9  & 81.3 & 18.6 & 18.4 \\
GradNorm\textcolor{red}{$^\dagger$}   & 78.4 & 21.4 & 20.4 & 78.4 & 5.8  & 5.6  & 78.4 & 18.5 & 18.4 \\
\midrule
LNL\textcolor{red}{$^*$}        & 69.5 & 12.9 & 11.1 & 73.5 & 4.5  & 4.3  & 74.8 & 15.4 & 15.1 \\
UFaTE\textcolor{red}{$^*$}      & 67.3 & 11.1 & 10.9 & 77.4 & 3.3  & 4.8  & 77.8 & 14.3 & 13.3 \\
\midrule
MGDA-Mean\textcolor{red}{$^{\dagger*}$}  & 62.0 & 16.5 & 15.1 & 74.5 & 3.7  & 3.6  & 72.8 & 13.4 & 13.1 \\
MGDA-UFaTE\textcolor{red}{$^{\dagger*}$} & 69.6 & 12.5 & 11.2 & 75.4 & 3.9  & 3.5  & 74.7 & 14.9 & 13.7 \\
\midrule
\rowcolor{gray!20} 
\textbf{FairMT} & \textbf{81.6} & \textbf{8.8} & \textbf{7.5} & \textbf{81.8} & \textbf{2.3} & \textbf{1.8} & \textbf{81.8} & \textbf{8.1} & \textbf{6.8} \\
\bottomrule
\end{tabular}}

\resizebox{\linewidth}{!}{
\setlength{\tabcolsep}{2pt}
\renewcommand{\arraystretch}{0.7}
\begin{tabular}{l|cccc|cccc|cccc}
\toprule
\multicolumn{13}{c}{\textbf{AffectNet-VA (Regression)}} \\
\midrule
\multirow{2}{*}{Method} & \multirow{2}{*}{CCC \textcolor{red}{$\uparrow$}} 
& \multicolumn{3}{c|}{Age}  & \multirow{2}{*}{CCC \textcolor{red}{$\uparrow$}}
& \multicolumn{3}{c|}{Gender}  & \multirow{2}{*}{CCC \textcolor{red}{$\uparrow$}}
& \multicolumn{3}{c}{Race} \\
\cmidrule(lr){3-5} \cmidrule(lr){7-9} \cmidrule(lr){11-13}
&  &  KS \textcolor{blue}{\textdownarrow}  & CSP \textcolor{blue}{\textdownarrow}  & EOD \textcolor{blue}{\textdownarrow} 
&   &  KS \textcolor{blue}{\textdownarrow}  & CSP \textcolor{blue}{\textdownarrow}  & EOD \textcolor{blue}{\textdownarrow} 
&   &  KS \textcolor{blue}{\textdownarrow}  & CSP \textcolor{blue}{\textdownarrow}  & EOD \textcolor{blue}{\textdownarrow} \\
\midrule
MGDA\textcolor{red}{$^\dagger$}       & \textbf{69.3} & 19.1 & 14.0 & 13.0 & 69.3 & 16.5 & 7.4 & 5.5 & 69.2 & 17.1 & 13.3 & 15.7 \\
GradNorm\textcolor{red}{$^\dagger$}   & 67.2 & 17.8 & 14.4 & 13.1 & 67.5 & 17.7 & 7.1 & 6.7 & 67.3 & 18.2 & 13.0 & 14.1 \\
\midrule
LNL\textcolor{red}{$^*$}        & 60.9 & 16.3 & 12.3 & 11.3 & 63.6 & 14.9 & 6.9 & 5.1 & 64.3 & 12.6 & 8.1  & 9.6  \\
UFaTE\textcolor{red}{$^*$}      & 58.5 & 14.3 & 12.9 & 8.6  & 63.0 & 13.1 & 6.1 & 5.3 & 63.8 & 13.1 & 7.9  & 10.9 \\
\midrule
MGDA-Mean\textcolor{red}{$^{\dagger*}$}  & 59.0 & 18.5 & 13.8 & 11.5 & 64.1 & 14.9 & 5.5 & 4.7 & 62.1 & 16.6 & 11.9 & 13.6 \\
MGDA-UFaTE\textcolor{red}{$^{\dagger*}$} & 61.9 & 17.8 & 13.4 & 12.1 & 65.9 & 13.8 & 6.8 & 5.8 & 64.3 & 10.0 & 7.1  & 8.7  \\
\midrule
\rowcolor{gray!20} 
\textbf{FairMT} & \underline{68.9} & \textbf{13.9} & \textbf{7.1} & \textbf{9.6} 
& \textbf{69.6} & \textbf{10.7} & \textbf{3.7} & \textbf{2.2} 
& \textbf{69.7} & \textbf{6.2} & \textbf{3.3} & \textbf{4.9} \\
\bottomrule
\end{tabular}}

\vspace{1em}
\noindent\textbf{\footnotesize Toxicity \& Social Bias Analysis (in \%)}
\resizebox{\linewidth}{!}{
\setlength{\tabcolsep}{7pt}
\renewcommand{\arraystretch}{0.3}
\begin{tabular}{l|ccc|ccc|ccc}
\toprule
\multicolumn{10}{c}{\textbf{CiviComment (Detection Task)}} \\
\midrule
\multirow{2}{*}{Method} & \multirow{2}{*}{Acc \textcolor{red}{$\uparrow$}} & \multicolumn{2}{c|}{LGBTQ+} & \multirow{2}{*}{Acc \textcolor{red}{$\uparrow$}}& \multicolumn{2}{c|}{Race}& \multirow{2}{*}{Acc \textcolor{red}{$\uparrow$}} & \multicolumn{2}{c}{Religion} \\
\cmidrule(lr){3-4} \cmidrule(lr){6-7} \cmidrule(lr){9-10}
 &  & EOD \textcolor{blue}{\textdownarrow} & EO \textcolor{blue}{\textdownarrow} &  & EOD \textcolor{blue}{\textdownarrow} & EO \textcolor{blue}{\textdownarrow} &  & EOD \textcolor{blue}{\textdownarrow} & EO \textcolor{blue}{\textdownarrow} \\
\midrule
MGDA\textcolor{red}{$^\dagger$}       & 87.1 & 14.3 & 14.1 & 81.6 & 22.9 & 22.2 & \textbf{84.1} & 28.1 & 27.9 \\
GradNorm\textcolor{red}{$^\dagger$}   & 86.2 & 13.1 & 12.8 & 78.7 & 23.2 & 23.3 & 83.1 & 29.0 & 29.3 \\
\midrule
LNL\textcolor{red}{$^*$}        & 82.6 & 7.6  & 7.6  & 74.1 & 14.2 & 14.7 & 77.6 & 17.7 & 19.7 \\
UFaTE\textcolor{red}{$^*$}      & 80.9 & 6.2  & 6.1  & 65.9 & 14.0 & 13.4 & 73.4 & 14.8 & 14.8 \\
\midrule
MGDA-Mean\textcolor{red}{$^{\dagger*}$}  & 81.9 & 11.8 & 11.9 & 74.4 & 17.8 & 16.5 & 75.1 & 15.7 & 14.8 \\
MGDA-UFaTE\textcolor{red}{$^{\dagger*}$} & 82.8 & 9.9  & 9.5  & 75.3 & 16.5 & 16.0 & 76.7 & 13.1 & 13.2 \\
\midrule
\rowcolor{gray!20} 
\textbf{FairMT} & \textbf{87.7} & \textbf{5.5} & \textbf{4.5} 
& \textbf{81.9} & \textbf{10.1} & \textbf{9.9} 
& \underline{83.8} & \textbf{6.7} & \textbf{6.0} \\
\bottomrule
\end{tabular}}

\resizebox{\linewidth}{!}{
\setlength{\tabcolsep}{2pt}
\renewcommand{\arraystretch}{0.6}
\begin{tabular}{l|cccc|cccc|cccc}
\toprule
\multicolumn{13}{c}{\textbf{CiviComment (Toxicity Regression)}} \\
\midrule
\multirow{2}{*}{Method} & \multirow{2}{*}{CCC \textcolor{red}{$\uparrow$}}
& \multicolumn{3}{c|}{LGBTQ+} & \multirow{2}{*}{CCC \textcolor{red}{$\uparrow$}}
& \multicolumn{3}{c|}{Race} & \multirow{2}{*}{CCC \textcolor{red}{$\uparrow$}}
& \multicolumn{3}{c}{Religion} \\
\cmidrule(lr){3-5} \cmidrule(lr){7-9} \cmidrule(lr){11-13}
&   &  KS \textcolor{blue}{\textdownarrow}  & CSP \textcolor{blue}{\textdownarrow}  & EOD \textcolor{blue}{\textdownarrow} 
&  &  KS \textcolor{blue}{\textdownarrow}  & CSP \textcolor{blue}{\textdownarrow}  & EOD \textcolor{blue}{\textdownarrow} 
&  &  KS \textcolor{blue}{\textdownarrow}  & CSP \textcolor{blue}{\textdownarrow}  & EOD \textcolor{blue}{\textdownarrow} \\
\midrule
MGDA\textcolor{red}{$^\dagger$}       & 73.4 & 34.9 & 12.5 & 27.4 & 69.2 & 42.7 & 18.2 & 38.9 & \textbf{72.0} & 39.8 & 15.8 & 35.3 \\
GradNorm\textcolor{red}{$^\dagger$}   & 71.5 & 39.5 & 14.9 & 26.8 & 68.4 & 41.1 & 18.8 & 35.6 & 69.4 & 39.9 & 22.3 & 38.3 \\
\midrule
LNL\textcolor{red}{$^*$}        & 67.7 & 28.3 & 9.7  & 23.9 & 65.3 & 32.9 & 15.1 & 27.9 & 66.1 & 29.9 & 9.5  & 21.6 \\
UFaTE\textcolor{red}{$^*$}      & 65.9 & 25.5 & 7.3  & 21.7 & 62.6 & 27.8 & 15.2 & 26.1 & 63.7 & 25.4 & 12.8 & 24.3 \\
\midrule
MGDA-Mean\textcolor{red}{$^{\dagger*}$}  & 67.6 & 31.1 & 11.2 & 24.3 & 63.7 & 36.7 & 16.2 & 31.9 & 64.4 & 33.9 & 12.3 & 31.3 \\
MGDA-UFaTE\textcolor{red}{$^{\dagger*}$} & 68.9 & 24.9 & 6.0  & 22.8 & 64.1 & 29.2 & 14.5 & 27.3 & 66.5 & 28.4 & 11.8 & 23.3 \\
\midrule
\rowcolor{gray!20} 
\textbf{FairMT} & \textbf{73.7} & \textbf{21.7} & \textbf{4.9} & \textbf{19.5} 
& \textbf{69.3} & \textbf{23.9} & \textbf{10.2} & \textbf{21.1} 
& \underline{71.1} & \textbf{20.7} & \textbf{7.2} & \textbf{19.9} \\
\bottomrule
\end{tabular}}
\vspace{-1em}
\end{table}

\begin{table*}[htbp]
\caption{Benchmark setups for fair multi-task learning across modalities, task regimes, supervision types, and sensitive groups.}
\label{tab:datasets_tasks}
\centering
\small
\resizebox{\linewidth}{!}{
\begin{tabular}{llllllll}
\toprule
\textbf{Task Name} & \textbf{\#Tasks} & \textbf{Modality} & \textbf{Task Regime} & \textbf{Supervision}& \textbf{Datasets} & \textbf{Task Types}  & \textbf{Sensitive Groups} \\
\midrule

\textbf{Attribute Detection} 
& 29
& Visual 
& Homogeneous
& Full labels
& CelebA\cite{liu2015faceattributes}
& \textbf{Detection:} 29 Facial Attributes
& \textbf{2 Age} \{Young, Old\}, \textbf{2 Gender} \{Male, Female\} \\
\midrule

\textbf{Affective Analysis} 
& 17
& Visual 
& Heterogeneous
& Partial labels
& \begin{tabular}{@{}l@{}}
AffectNet\cite{mollahosseini2017affectnet}\\
AffectNet-VA\cite{mollahosseini2017affectnet}\\
EmotioNet\cite{fabian2016emotionet}
\end{tabular}
& \begin{tabular}{@{}l@{}}
\textbf{Classification:} 7 Expressions\\
\textbf{Regression:} Valence \& Arousal \\
\textbf{Detection:} 8 Action-Unit Detection
\end{tabular}

& \begin{tabular}{@{}l@{}}
\textbf{7 Age}: \{00--09, 10--19, 20--29, 30--39,40--49, 50--59, 60+\}\\
\textbf{2 Gender}: \{Male, Female\}\\
\textbf{4 Race}: \{Asian, Black, Indian, White\}
\end{tabular} \\
\midrule
\textbf{Toxicity \& Social Bias Analysis} 
& 9
& Text
& Heterogeneous
& Full labels
& Civil Comments\cite{koh2021wilds}
& \begin{tabular}{@{}l@{}}
\textbf{Detection:} Rating, Identity Attack,\\
Insult, Obscene,  Likes, Disagree,\\
Funny, Wow, Sad\\
\textbf{Regression:}  Toxicity Value
\end{tabular}
& \begin{tabular}{@{}l@{}}
\textbf{2 Sexual Orientation}: \{LGBTQ+, non-LGBTQ+\}\\
\textbf{3 Religions}: \{Christian, Muslim, No-Religions\}\\
\textbf{5 Race}: \{Asian, Black,White, Latino, Other\}
\end{tabular} \\
\bottomrule
\end{tabular}
}
\vspace{-1.5em}
\end{table*}

\textbf{MTL Settings} We evaluate \textsc{FairMT} on three large-scale multi-task benchmarks spanning both vision and language modalities, mixing homogeneous and heterogeneous task structures, and covering both full and partial supervision. 
These settings jointly comprise detection, classification, and regression objectives, and require fairness assessment across diverse demographic axes including age, gender, race, religion, and sexual orientation, as summarized in Table~\ref{tab:datasets_tasks}. \textbf{\textit{Please Refer to Appendix \textcolor{red}{D} for comprehensive descriptions of the datasets, implementation settings, evaluation metrics (Fairness \& Utility), and SOTA baselines.}}

\begin{figure}[htbp]
\centering
\includegraphics[width=1\linewidth]{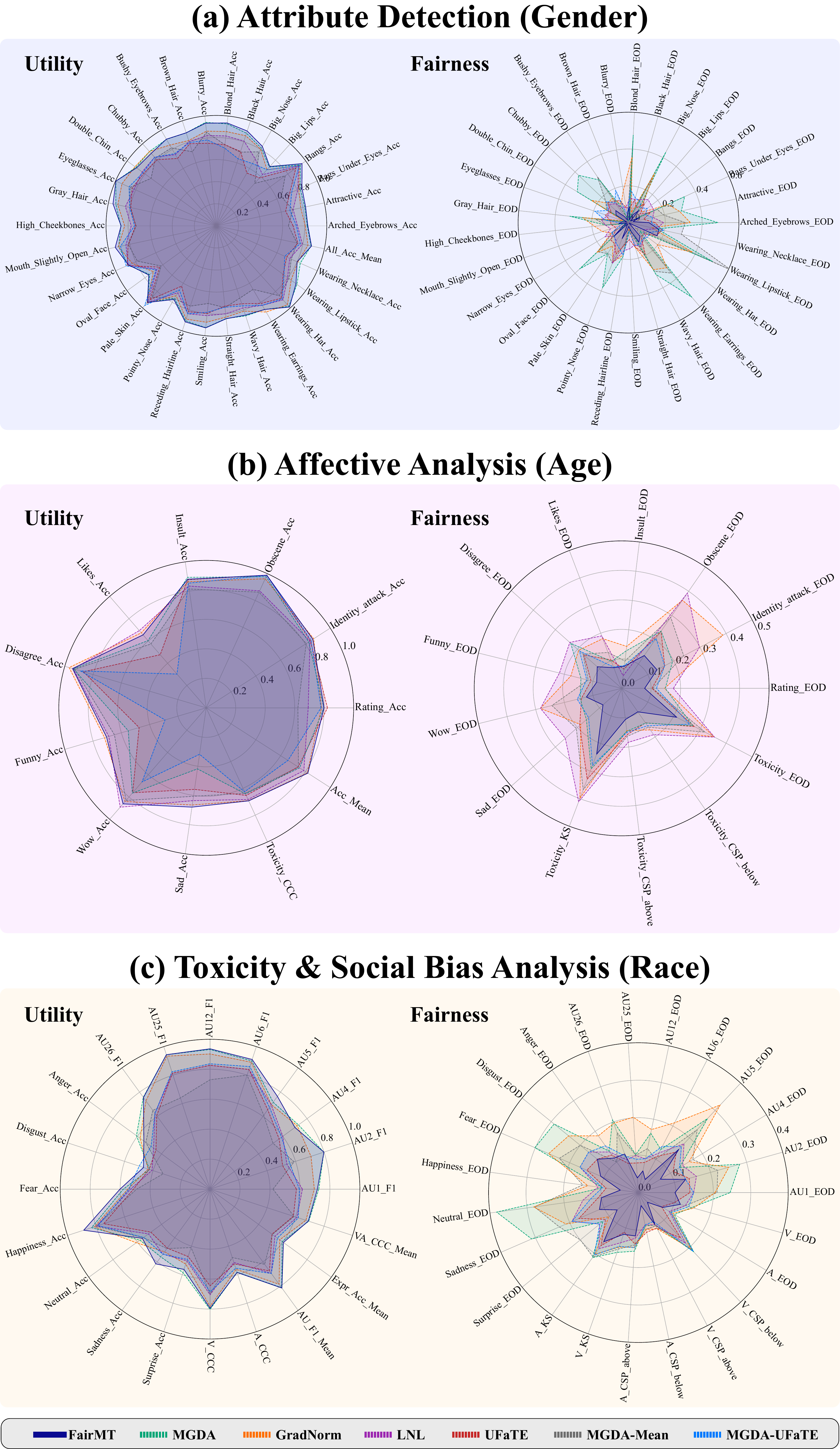}
\caption{
{\small
Comparison between \textbf{FairMT} and state-of-the-art baselines across three multi-task settings:
(a) Attribute Detection,
(b) Affective Analysis,
and (c) Toxicity \& Social Bias Analysis.
}
\vspace{-1em}
}
\label{fig:radar}
\end{figure}

\textbf{Comparison with state-of-the-art (SOTA).}
As shown in Figure~\ref{fig:radar} and Table~\ref{tab:sota}, \textsc{FairMT} consistently 
achieves superior fairness \emph{without incurring utility trade-off}, and even attains the 
best task performance across most benchmarks, modalities, and task compositions. 
We compare against representative SOTA baselines across three categories. 
For multi-task optimization, we include \textbf{MGDA}~\cite{sener2018multi} and 
\textbf{GradNorm}~\cite{chen2018gradnorm}. 
For fairness-oriented learning, we adopt \textbf{LNL}~\cite{kim2019learning} 
and \textbf{UFaTE}~\cite{dehdashtian2024utility}. 
Since there is no comparable approach, we further construct two hybrid variants: 
\textbf{MGDA-Mean}, which applies MGDA under a naïve averaged disparity constraint by enforcing 
Eq.~\eqref{eq:constrained_mtl} on mean task disparities, and \textbf{MGDA-UFaTE}, which 
integrates MGDA with the UFaTE regularizer to jointly optimize utility and fairness. 
Multi-task optimizers (MGDA, GradNorm) improve overall utility but yield noticeable fairness 
gaps, whereas fairness-driven methods (LNL, UFaTE) reduce demographic disparities at the 
cost of predictive accuracy. Hybrid designs such as MGDA-Mean and MGDA-UFaTE partially 
alleviate this tension but remain inferior to \textsc{FairMT} on fairness/utility.

\subsection{Ablation Study and Further Analysis}

\begin{figure}[htbp]
\centering
\includegraphics[width=1\linewidth]{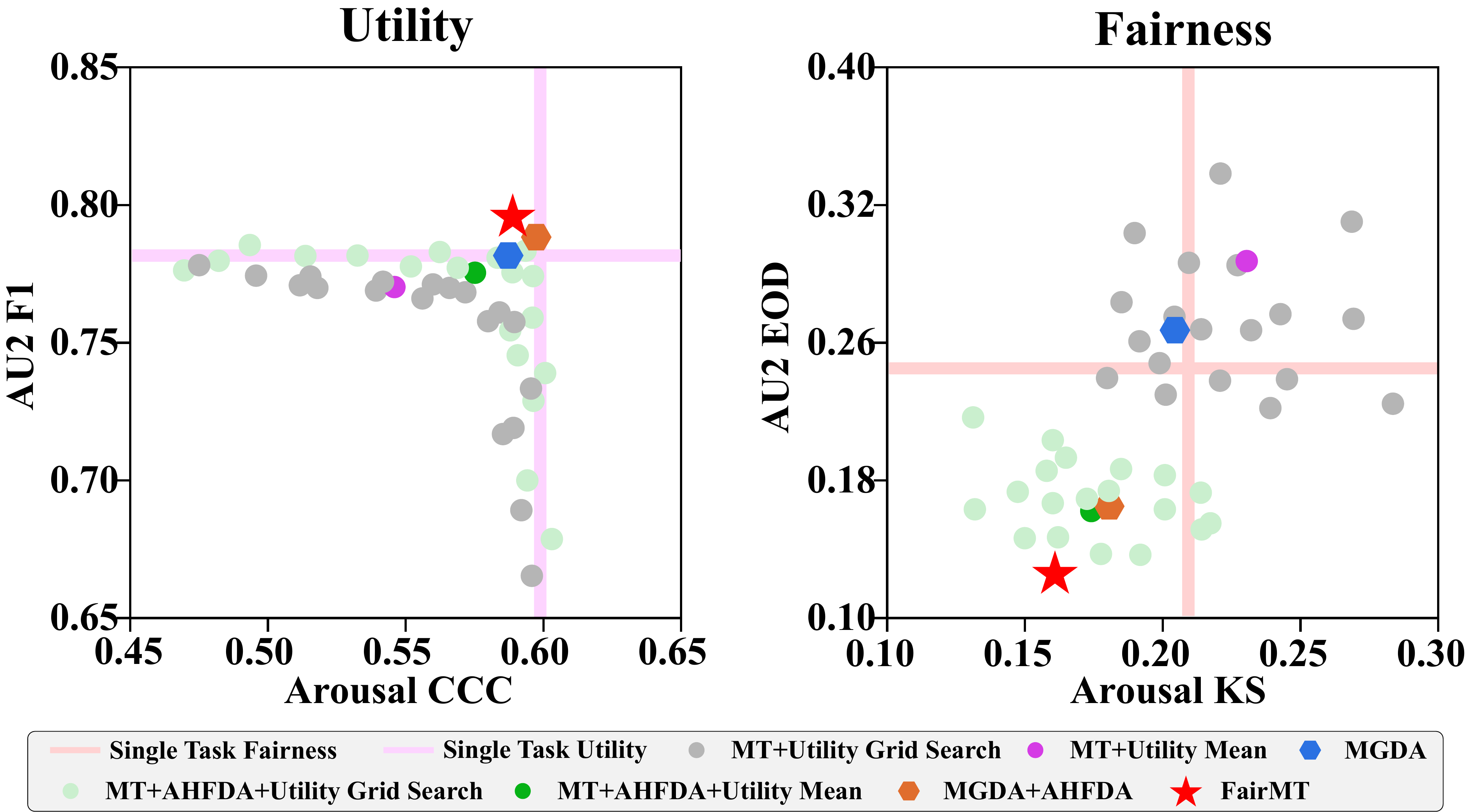}
\caption{
{\footnotesize
\textbf{Ablation on heterogeneous AU2--Arousal tasks.} 
We compare conventional MTL, utility-weighted baselines, MGDA, and their AHFDA-enhanced variants on both utility (left) and fairness (right). 
}}
\vspace{-1em}
\label{fig:ablation}
\end{figure}

We comparing conventional MTL, utility-weighted baselines, MGDA, and their AHFDA-augmented variants. As shown in Fig.~\ref{fig:ablation}, \textsc{FairMT} attains a strictly better utility–fairness Pareto point. Adding AHFDA improves MT by lifting disadvantaged groups without degrading advantaged ones, and similarly strengthens MGDA, though its performance remains limited by MGDA’s geometry-agnostic surrogate. Once MGDA’s proxy is replaced with our head-aware formulation, both utility and fairness improve consistently, underscoring the importance of modeling head geometry for aligned fairness-aware MTL optimization \textbf{(Please see Appendix~\textcolor{red}{E} for more detailed performance and efficiency analyses)}.

\section{Conclusion}
We presented \textsc{FairMT}, a unified fairness-aware multi-task learning framework capable of handling heterogeneous tasks under partial supervision. By introducing the Asymmetric Heterogeneous Fairness Disparity Aggregation and a Pareto-guided optimization strategy, \textsc{FairMT} improves under-served groups without sacrificing overall task performance. Experiments across vision and language benchmarks show consistent gains in both utility and fairness across diverse datasets. This work illustrates that fairness and multi-task efficacy can be jointly optimized, opening avenues for extending \textsc{FairMT} to broader fairness notions and more complex real-world settings.

{
    \small
    \bibliographystyle{ieeenat_fullname}
    \bibliography{main}

@String(ICCV= {Int. Conf. Comput. Vis.})

@String(ICCV  = {ICCV})

@inproceedings{zafar2017fairness,
  title={Fairness constraints: Mechanisms for fair classification},
  author={Zafar, Muhammad Bilal and Valera, Isabel and Rogriguez, Manuel Gomez and Gummadi, Krishna P},
  booktitle={Artificial intelligence and statistics},
  pages={962--970},
  year={2017},
  organization={PMLR}
}

@inproceedings{menon2018cost,
  title={The cost of fairness in binary classification},
  author={Menon, Aditya Krishna and Williamson, Robert C},
  booktitle={Conference on Fairness, accountability and transparency},
  pages={107--118},
  year={2018},
  organization={PMLR}
}

@article{zafar2019fairness,
  title={Fairness constraints: A flexible approach for fair classification},
  author={Zafar, Muhammad Bilal and Valera, Isabel and Gomez-Rodriguez, Manuel and Gummadi, Krishna P},
  journal={Journal of Machine Learning Research},
  volume={20},
  number={75},
  pages={1--42},
  year={2019}
}

@article{mehrabi2021survey,
  title={A survey on bias and fairness in machine learning},
  author={Mehrabi, Ninareh and Morstatter, Fred and Saxena, Nripsuta and Lerman, Kristina and Galstyan, Aram},
  journal={ACM computing surveys (CSUR)},
  volume={54},
  number={6},
  pages={1--35},
  year={2021},
  publisher={ACM New York, NY, USA}
}

@article{pessach2022review,
  title={A review on fairness in machine learning},
  author={Pessach, Dana and Shmueli, Erez},
  journal={ACM Computing Surveys (CSUR)},
  volume={55},
  number={3},
  pages={1--44},
  year={2022},
  publisher={ACM New York, NY}
}

@article{soremekun2022software,
  title={Software fairness: An analysis and survey},
  author={Soremekun, Ezekiel and Papadakis, Mike and Cordy, Maxime and Le Traon, Yves},
  journal={ACM Computing Surveys},
  year={2022},
  publisher={ACM New York, NY}
}

@article{lin2019pareto,
  title={Pareto multi-task learning},
  author={Lin, Xi and Zhen, Hui-Ling and Li, Zhenhua and Zhang, Qing-Fu and Kwong, Sam},
  journal={Advances in neural information processing systems},
  volume={32},
  year={2019}
}

@inproceedings{mahapatra2020multi,
  title={Multi-task learning with user preferences: Gradient descent with controlled ascent in pareto optimization},
  author={Mahapatra, Debabrata and Rajan, Vaibhav},
  booktitle={International Conference on Machine Learning},
  pages={6597--6607},
  year={2020},
  organization={PMLR}
}

@article{yu2020gradient,
  title={Gradient surgery for multi-task learning},
  author={Yu, Tianhe and Kumar, Saurabh and Gupta, Abhishek and Levine, Sergey and Hausman, Karol and Finn, Chelsea},
  journal={Advances in neural information processing systems},
  volume={33},
  pages={5824--5836},
  year={2020}
}

@article{berk2017convex,
  title={A convex framework for fair regression},
  author={Berk, Richard and Heidari, Hoda and Jabbari, Shahin and Joseph, Matthew and Kearns, Michael and Morgenstern, Jamie and Neel, Seth and Roth, Aaron},
  journal={arXiv preprint arXiv:1706.02409},
  year={2017}
}

@article{chzhen2020fair,
  title={Fair regression via plug-in estimator and recalibration with statistical guarantees},
  author={Chzhen, Evgenii and Denis, Christophe and Hebiri, Mohamed and Oneto, Luca and Pontil, Massimiliano},
  journal={Advances in Neural Information Processing Systems},
  volume={33},
  pages={19137--19148},
  year={2020}
}

@inproceedings{agarwal2018reductions,
  title={A reductions approach to fair classification},
  author={Agarwal, Alekh and Beygelzimer, Alina and Dud{\'\i}k, Miroslav and Langford, John and Wallach, Hanna},
  booktitle={International conference on machine learning},
  pages={60--69},
  year={2018},
  organization={PMLR}
}

@article{desideri2012multiple,
  title={Multiple-gradient descent algorithm (MGDA) for multiobjective optimization},
  author={D{\'e}sid{\'e}ri, Jean-Antoine},
  journal={Comptes Rendus Mathematique},
  volume={350},
  number={5-6},
  pages={313--318},
  year={2012},
  publisher={Elsevier}
}

@article{sener2018multi,
  title={Multi-task learning as multi-objective optimization},
  author={Sener, Ozan and Koltun, Vladlen},
  journal={Advances in neural information processing systems},
  volume={31},
  year={2018}
}

@article{wolfe1976finding,
  title={Finding the nearest point in a polytope},
  author={Wolfe, Philip},
  journal={Mathematical Programming},
  volume={11},
  number={1},
  pages={128--149},
  year={1976},
  publisher={Springer}
}

@article{sekitani1993recursive,
  title={A recursive algorithm for finding the minimum norm point in a polytope and a pair of closest points in two polytopes},
  author={Sekitani, Kazuyuki and Yamamoto, Yoshitsugu},
  journal={Mathematical Programming},
  volume={61},
  number={1},
  pages={233--249},
  year={1993},
  publisher={Springer}
}

@inproceedings{garg2020fairness,
  title={Fairness metrics: A comparative analysis},
  author={Garg, Pratyush and Villasenor, John and Foggo, Virginia},
  booktitle={2020 IEEE international conference on big data (Big Data)},
  pages={3662--3666},
  year={2020},
  organization={IEEE}
}

@inproceedings{tang2022attainability,
  title={Attainability and optimality: The equalized odds fairness revisited},
  author={Tang, Zeyu and Zhang, Kun},
  booktitle={Conference on Causal Learning and Reasoning},
  pages={754--786},
  year={2022},
  organization={PMLR}
}

@inproceedings{zhong2024intrinsic,
  title={Intrinsic fairness-accuracy tradeoffs under equalized odds},
  author={Zhong, Meiyu and Tandon, Ravi},
  booktitle={2024 IEEE International Symposium on Information Theory (ISIT)},
  pages={220--225},
  year={2024},
  organization={IEEE}
}

@article{chzhen2020fair1,
  title={Fair regression via plug-in estimator and recalibration with statistical guarantees},
  author={Chzhen, Evgenii and Denis, Christophe and Hebiri, Mohamed and Oneto, Luca and Pontil, Massimiliano},
  journal={Advances in Neural Information Processing Systems},
  volume={33},
  pages={19137--19148},
  year={2020}
}

@inproceedings{wei2023mean,
  title={Mean parity fair regression in rkhs},
  author={Wei, Shaokui and Liu, Jiayin and Li, Bing and Zha, Hongyuan},
  booktitle={International conference on artificial intelligence and statistics},
  pages={4602--4628},
  year={2023},
  organization={PMLR}
}

@inproceedings{koh2021wilds,
  title={Wilds: A benchmark of in-the-wild distribution shifts},
  author={Koh, Pang Wei and Sagawa, Shiori and Marklund, Henrik and Xie, Sang Michael and Zhang, Marvin and Balsubramani, Akshay and Hu, Weihua and Yasunaga, Michihiro and Phillips, Richard Lanas and Gao, Irena and others},
  booktitle={International conference on machine learning},
  pages={5637--5664},
  year={2021},
  organization={PMLR}
}

@article{mollahosseini2017affectnet,
  title={Affectnet: A database for facial expression, valence, and arousal computing in the wild},
  author={Mollahosseini, Ali and Hasani, Behzad and Mahoor, Mohammad H},
  journal={IEEE Transactions on Affective Computing},
  volume={10},
  number={1},
  pages={18--31},
  year={2017},
  publisher={IEEE}
}

@inproceedings{fabian2016emotionet,
  title={Emotionet: An accurate, real-time algorithm for the automatic annotation of a million facial expressions in the wild},
  author={Fabian Benitez-Quiroz, C and Srinivasan, Ramprakash and Martinez, Aleix M},
  booktitle={Proceedings of the IEEE conference on computer vision and pattern recognition},
  pages={5562--5570},
  year={2016}
}

@inproceedings{liu2015faceattributes,
  title = {Deep Learning Face Attributes in the Wild},
  author = {Liu, Ziwei and Luo, Ping and Wang, Xiaogang and Tang, Xiaoou},
  booktitle = {Proceedings of International Conference on Computer Vision (ICCV)},
  month = {December},
  year = {2015} 
}

@inproceedings{chen2018gradnorm,
  title={Gradnorm: Gradient normalization for adaptive loss balancing in deep multitask networks},
  author={Chen, Zhao and Badrinarayanan, Vijay and Lee, Chen-Yu and Rabinovich, Andrew},
  booktitle={International conference on machine learning},
  pages={794--803},
  year={2018},
  organization={PMLR}
}

@inproceedings{kim2019learning,
  title={Learning not to learn: Training deep neural networks with biased data},
  author={Kim, Byungju and Kim, Hyunwoo and Kim, Kyungsu and Kim, Sungjin and Kim, Junmo},
  booktitle={Proceedings of the IEEE/CVF conference on computer vision and pattern recognition},
  pages={9012--9020},
  year={2019}
}

@inproceedings{dehdashtian2024utility,
  title={Utility-fairness trade-offs and how to find them},
  author={Dehdashtian, Sepehr and Sadeghi, Bashir and Boddeti, Vishnu Naresh},
  booktitle={Proceedings of the IEEE/CVF Conference on Computer Vision and Pattern Recognition},
  pages={12037--12046},
  year={2024}
}

@inproceedings{madras2018learning,
  title={Learning adversarially fair and transferable representations},
  author={Madras, David and Creager, Elliot and Pitassi, Toniann and Zemel, Richard},
  booktitle={International Conference on Machine Learning},
  pages={3384--3393},
  year={2018},
  organization={PMLR}
}

@inproceedings{zhang2018mitigating,
  title={Mitigating unwanted biases with adversarial learning},
  author={Zhang, Brian Hu and Lemoine, Blake and Mitchell, Margaret},
  booktitle={Proceedings of the 2018 AAAI/ACM Conference on AI, Ethics, and Society},
  pages={335--340},
  year={2018}
}

@inproceedings{quadrianto2019discovering,
  title={Discovering fair representations in the data domain},
  author={Quadrianto, Novi and Sharmanska, Viktoriia and Thomas, Oliver},
  booktitle={Proceedings of the IEEE/CVF conference on computer vision and pattern recognition},
  pages={8227--8236},
  year={2019}
}

@inproceedings{agarwal2019fair,
  title={Fair regression: Quantitative definitions and reduction-based algorithms},
  author={Agarwal, Alekh and Dud{\'\i}k, Miroslav and Wu, Zhiwei Steven},
  booktitle={International conference on machine learning},
  pages={120--129},
  year={2019},
  organization={PMLR}
}
}

\end{document}